\documentclass[letterpaper, 10 pt, conference]{ieeeconf}  

\IEEEoverridecommandlockouts                              

\usepackage{graphics} 
\usepackage{mathtools}
\usepackage{times} 
\usepackage{amsmath} 
\usepackage{amssymb}  
\usepackage{xcolor}
\usepackage[utf8]{inputenc}
\usepackage{comment}
\usepackage{algorithm,algorithmic}
\usepackage{amsthm}
\usepackage{balance}
\usepackage{hyperref}

\usepackage{multirow}
\usepackage{graphicx}

\usepackage{tikz}

\DeclareMathAlphabet\mathbfcal{OMS}{cmsy}{b}{n}

\usepackage{cancel}
\theoremstyle{plain}
\newtheorem{remark}{Remark}

\newcommand{\q}{\boldsymbol{q}}
\newcommand{\R}{\mathbb{R}}

\newcommand{\y}{\boldsymbol{y}}

\newcommand{\ti}{\text{i}}

\DeclareMathAlphabet{\mathcal}{OMS}{cmsy}{m}{n}

\title{\LARGE \bf
RoMoCo: \underline{Ro}botic \underline{Mo}tion \underline{Co}ntrol Toolbox for Reduced-Order Model-Based Locomotion on Bipedal and Humanoid Robots
}

\author{Min Dai and Aaron D. Ames
\thanks{ The authors are with the Department of Mechanical and Civil Engineering, California Institute of Technology, Pasadena, CA 91125 USA.
        {\tt\small \{mdai,ames\}@caltech.edu}.
        }
}

\begin{document}

\maketitle
\thispagestyle{empty}
\pagestyle{empty}

\begin{abstract}

We present RoMoCo, an open-source C++ toolbox for the synthesis and evaluation of reduced-order model-based planners and whole-body controllers for bipedal and humanoid robots. RoMoCo's modular architecture unifies state-of-the-art planners and whole-body locomotion controllers under a consistent API, enabling rapid prototyping and reproducible benchmarking. By leveraging reduced-order models for platform-agnostic gait generation, RoMoCo enables flexible controller design across diverse robots.
We demonstrate its versatility and performance through extensive simulations on the Cassie, Unitree H1, and G1 robots, and validate its real-world efficacy with hardware experiments on the Cassie and G1 humanoids.
\end{abstract}


\section{Introduction}

Bipedal locomotion remains one of the central challenges in robotics, given bipeds' high-dimensional, non-linear, hybrid, and underactuated nature. A widely adopted strategy to address this complexity is the use of reduced-order models (ROM) that capture the essential dynamics of walking while abstracting full-body details. These models, including the linear inverted pendulum (LIP)~\cite{kajita_3d_2001} and its variants~\cite{gong_angular_2021, xiong_3-d_2022}, have enabled the design of theoretically grounded and robust locomotion controllers. However, despite their success in research, deploying ROM-based planners and integrating them with whole-body controllers (WBC) remains a significant challenge, requiring expertise in contact and state estimation, robot kinematics and dynamics, and nonlinear control. 


In recent years, there has been a surge of interest in learning-based locomotion, largely fueled by the release of frameworks such as IsaacLab \cite{mittal2023orbit} and IsaacGym \cite{makoviychuk2021isaac}, which provide scalable reinforcement learning (RL) environments and simulation infrastructure. These platforms, combined with open-source RL algorithms implementations such as RSL-RL \cite{rudin_learning_2022}, have lowered the barrier to entry for training locomotion policies at scale, enabling impressive demonstrations of locomotion in simulation and on hardware.  

In contrast to learning-based approaches, model-based methods, though more interpretable, computationally efficient, and theoretically grounded, lack equivalent open-source support. Researchers typically face steep implementation hurdles, not only in developing reduced-order planners but also in integrating them with whole-body controllers and simulators. To date, only isolated efforts exist, such as the ALIP controller for Cassie \cite{gong_alip_code} implemented in Simulink Real-Time, which, while effective, is tied to a specific robot platform and simulation ecosystem, preventing its generalization. This lack of a standardized, extensible framework limits reproducibility, comparative benchmarking, and rapid prototyping across diverse robotic platforms.

\begin{figure}[t]
\centering\includegraphics[width=\linewidth]{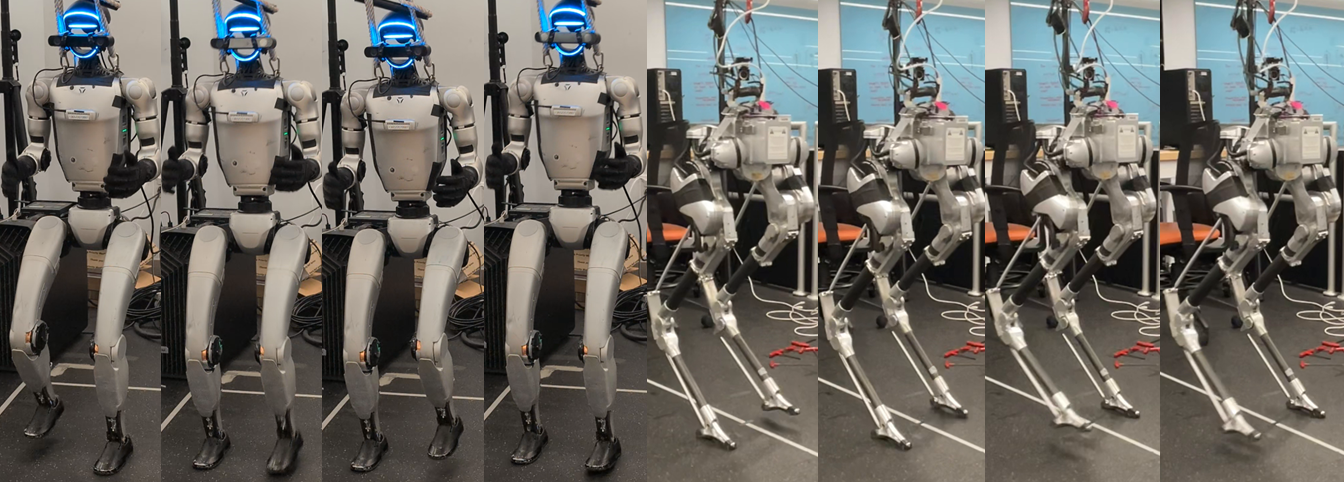}
  \caption{G1 and Cassie stepping-in-place using the RoMoCo toolbox. }
  \vspace{-5mm}
  \label{fig::gaittile}
\end{figure}

\begin{figure*}[t]
  \vspace{2mm}
\centering\includegraphics[width=\textwidth]{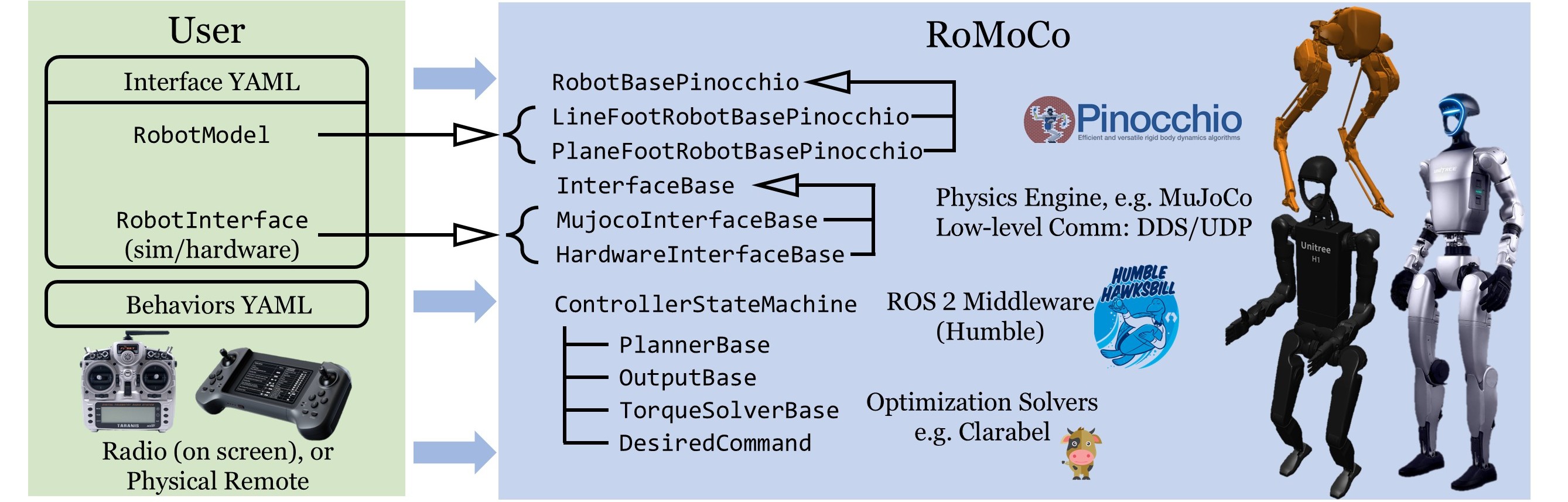}
  \caption{The \texttt{RoMoCo} library is organized around modular components: core robot abstractions are built on Pinocchio, interfaces support both simulation and hardware, and planners, outputs, and torque solvers are coordinated by a controller state machine. Desired behaviors are specified through software or physical control, while ROS~2 middleware provides communication.}
  \vspace{-4mm}
  \label{fig::system_overview}
\end{figure*}
A number of open-source libraries provide essential components for the planning and control of robots. FROST \cite{hereid_frost_2017} supports trajectory optimization for bipedal robots, though it emphasizes offline trajectory generation rather than real-time controllers. Drake \cite{drake} offers a versatile platform for trajectory optimization and dynamics simulation, but its focus remains primarily on manipulation. OCS2 \cite{OCS2} implements efficient optimal control solvers for switched systems, but it requires significant integration effort to connect with whole-body controllers or reduced-order abstractions. Other optimal control tools, such as OpenSoT~\cite{opensot:icra15}, Crocoddyl~\cite{mastalli_crocoddyl_2020},  and TSID~\cite{adelprete:jnrh:2016}, operate largely in isolation from dynamic locomotion planning. Collectively, these libraries form a rich ecosystem of building blocks; yet, researchers still assemble them manually, as there remains no unified framework that seamlessly couples reduced-order planning with whole-body control in a reproducible, extensible manner.


To address these challenges, we introduce RoMoCo—an open-source software toolbox designed to unify the development, evaluation, and deployment of ROM-based planners and WBC algorithms for bipedal and humanoid locomotion. The key contributions of this work are:
\begin{itemize}
    \item A unified mathematical formulation of popular LIP-based planners (ALIP, H-LIP, MLIP, DCM), enabling their modular implementation and direct comparison.
    \item A modular software architecture that decouples planners, output mappings, whole-body controllers, and robot interfaces, allowing for rapid prototyping across different hardware.
    \item An open-source library with integrated MuJoCo~\cite{todorov_mujoco_2012} simulation, demonstrated hardware deployments on multiple bipedal and humanoid platforms, and a fully available anonymized repository\footnote{\url{https://anonymous.4open.science/r/RoMoCo-6E85}} for review purposes.\footnote{The repository will be linked to a permanent public release upon acceptance.}
    \item A comparative analysis of different ROM planners and whole-body controllers on the Cassie and Unitree G1, offering insights into their performance trade-offs.
\end{itemize}




\section{System Overview}

RoMoCo is a locomotion control library with a modular architecture (Fig.~\ref{fig::system_overview}). As summarized in Table~\ref{tab:romoco_packages} (\texttt{romoco\_} prefix omitted in table for brevity), the \texttt{romoco\_core} package provides robot abstractions and Pinocchio-based kinematics and dynamics~\cite{carpentier_pinocchio_2019}, ensuring that higher-level modules can be reused across simulation and hardware. Building on this base, specialized packages implement reduced-order planners, output embeddings, torque solvers, and a controller state machine, while supporting packages handle MuJoCo simulation, utilities, and ROS~2 communication~\cite{ros2_2022}. 

\begin{table}[h]
\vspace{-3mm}
\centering
\caption{Summary of RoMoCo packages}
\label{tab:romoco_packages}
\scriptsize
\begin{tabular}{ll}
\hline
\textbf{Package} & \textbf{Function} \\
\hline
\texttt{core} & Robot Kinematics/dynamics, base classes \\
\texttt{planner} & ROM planners (ALIP, H-LIP, MLIP, DCM) \\
\texttt{output} & ROM trajectory embedding for walking, standing \\
\texttt{control} & Torque solvers, first and second order methods \\
\texttt{mujoco} & MuJoCo simulation interface \\
\texttt{state\_machine} & Locomotion phase management \\
\texttt{utils} & Utilities (Bezier, YAML, geometry) \\
\texttt{msgs} & ROS~2 messages \\
\texttt{ros} & ROS~2 nodes \\
\texttt{screen\_radio} & Command interface \\
\hline
\end{tabular}
\vspace{-3mm}
\end{table}

\begin{figure}[t]
\centering\includegraphics[width=\linewidth]{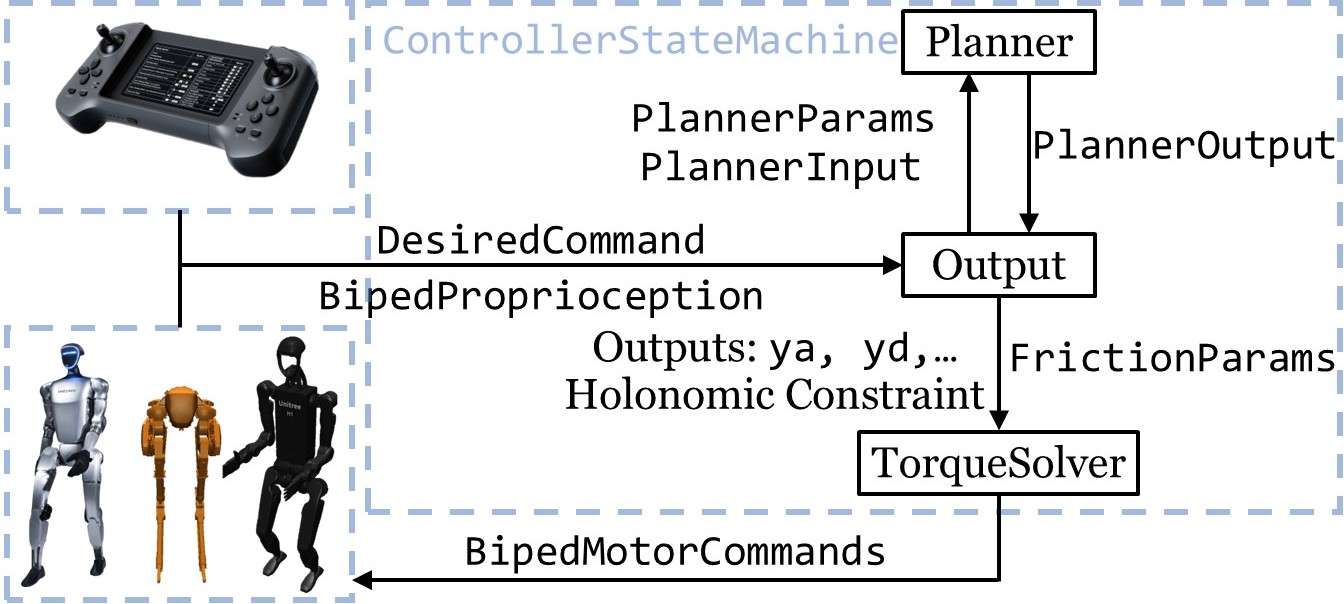}
  \vspace{-5mm}
  \caption{The \texttt{ControllerStateMachine} coordinates the ROM planner, output embedding, 
and torque solver, producing motor commands consistent with holonomic constraints.}
  \vspace{-7mm}
  \label{fig::controller_overview}
\end{figure}

The overall control pipeline follows the classical ROM-based locomotion structure (Fig.~\ref{fig::controller_overview}). First, a ROM planner generates high-level planner outputs such as foot placements. These discrete references are then mapped to whole-body task references (e.g., swing foot trajectories, torso orientations), which serve as inputs to a whole-body controller. The controller, such as operational space control or inverse kinematics, computes consistent motor commands that realize the planned behaviors while respecting the robot’s kinematic and dynamic constraints. Finally, the motor commands are applied to the simulated or real robot through the RoMoCo interface, closing the loop with sensor feedback. 

Integration of a new robot requires only minimal customization, as the pipeline itself remains unchanged. A user provides a robot description by implementing a \texttt{RobotModel} and a \texttt{RobotInterface} inherited from their base classes. The model encodes kinematics, dynamics, and joint indices, while the interface specifies how commands are applied in simulation or hardware. Additional configuration is specified in YAML files, which define robot parameters, planner and controller selection with parameters, and runtime command interfaces. This separation of concerns ensures that the provided planning and control modules can be deployed across different robot platforms with only lightweight modifications at the user's end.

\section{RoM based Locomotion Planners}
Bipedal locomotion is inherently a hybrid dynamical system with continuous-time dynamics during swing phases and discrete transitions at foot-ground impact~\cite{grizzle_models_2014}. While full-order robot dynamics are high-dimensional and nonlinear, much of the essential step-to-step (S2S) behavior can be captured by simplified, low-dimensional models. In this section, we review several widely used ROMs for locomotion planning and show how they can be expressed within a unified state-space framework. This perspective not only highlights structural similarities across models but also enables modular implementation within our planning and output embedding packages \texttt{romoco\_planner} and \texttt{romoco\_output}. For brevity, we present sagittal-plane formulations, though all models generalize to 3D walking; the coronal-plane implementations are available in the annotated codebase. The current release focuses on LIP-based foot-placement planners, including ALIP~\cite{gong_angular_2021}, H-LIP~\cite{xiong_3-d_2022}, 
MLIP~\cite{dai_multi-domain_2024}, and DCM~\cite{englsberger_three-dimensional_2013}. The RoMoCo framework is readily extensible to more advanced formulations beyond foot placement~\cite{gibson_terrain-aware_2021, dai_robust_2025}.
\begin{figure}[t]
\centering
\vspace{2mm}
  \includegraphics[width=\linewidth]{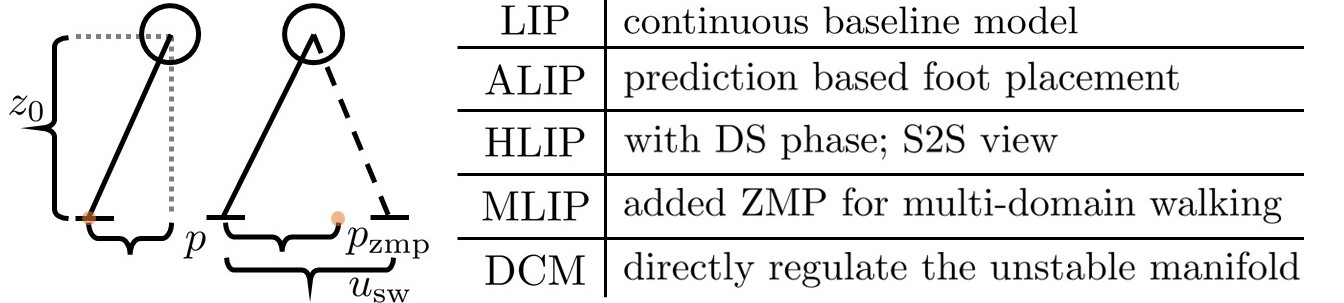}
  \caption{Visualization of LIP model parameters in single support (SS) and double support (DS). The table summarizes extensions, ALIP, H-LIP, MLIP, and DCM, highlighting their distinguishing features.}
  \vspace{-6mm}
  \label{fig::lip}
\end{figure}
We begin with the canonical LIP model, which serves as the foundation for all extensions. As shown in Fig.~\ref{fig::lip}, it includes a point mass moving at a constant CoM height $z_0$, supported by two massless, telescopic legs with point contacts. Here we consider flat ground, though the dynamics remain linear for sloped terrain~\cite{dai_bipedal_2022} and extend approximately to 3D under simplifying assumptions~\cite{gibson_terrain-aware_2021}. Neglecting ankle actuation, the sagittal-plane dynamics are given by
$\ddot{p} = \frac{g}{z_0} p \eqqcolon \lambda^2 p$, 
where $p$ is the horizontal CoM position relative to the stance pivot and $\lambda = \sqrt{\frac{g}{z_0}}$. While the classic formulation uses $[p, \dot{p}]^\intercal$ as the state, we instead use the mass-normalized angular momentum about the stance pivot, $L$, as the second state. This choice better reflects the model's similarity to robots' underactuated dynamics and improves state estimation~\cite{gong_zero_2022}. Under this representation, the state-space dynamics and closed-form solution are:
\begin{align}\label{eq::LIPode}
\dot{\mathbf{x}} &= 
\frac{d}{dt}
    \begin{bmatrix} p\\L\end{bmatrix} = \begin{bmatrix} 0 & \frac{1}{z_0}\\g & 0\end{bmatrix}\begin{bmatrix} p\\L\end{bmatrix} \coloneqq A_\text{SS}\mathbf{x} ,\\
\begin{bmatrix}
p(t) \\
L(t)
\end{bmatrix}
&=
\begin{bmatrix}
\cosh(\lambda t) & \frac{1}{ z_0 \lambda} \sinh( \lambda t) \\
 z_0 \lambda \sinh(\lambda t) & \cosh(\lambda t)
\end{bmatrix}
\begin{bmatrix}
p(0) \\
L(0)
\end{bmatrix}.
\end{align}
At foot-strike, let $u_\text{sw}$ denote the horizontal swing-foot location relative to the current stance foot. With the input matrix $B_\Delta = [-1 \, 0]^\intercal$,  the stance-to-swing transition is represented by the discrete update given by:
\begin{equation}
\mathbf{x}^+=\mathbf{x}^- + B_\Delta u_\text{sw},
\end{equation}
where $(\cdot)^+$ and $(\cdot)^-$ are the pre- and post-impact states.
\subsection{Angular Momentum(A)-LIP}
The ALIP model~\cite{gong_angular_2021} extends the LIP by explicitly predicting the pre-impact angular momentum and solving for foot placement for momentum regulation. At the current time $t$, the pre-impact angular momentum at the next impact $T_k$ can be estimated from the closed-form LIP solution as
\begin{align*}
\scalebox{0.9}{$
    \hat{L}^-(T_k | t) = z_0\lambda \sinh (\lambda (T_k - t) ) p(t) + \cosh(\lambda (T_k - t)) L(t).
$} 
\end{align*}
Assuming a step duration $T = T_\text{SS}$, the predicted momentum for the next step ahead is
\begin{align*}
&\scalebox{0.85}{$
\hat{L}^-(T_{k+1}|t) = z_0\lambda \sinh (\lambda T ) \hat{p}^+(T_k) + \cosh(\lambda T) \hat{L}^+(T_k) 
$} \\
&\qquad\qquad\,\;\scalebox{0.85}{$ = z_0\lambda \sinh (\lambda T ) (\hat{p}^-(T_k|t) - u_\text{sw}) + \cosh(\lambda T) \hat{L}^-(T_k|t),
$}     
\end{align*}
where $\hat{p}^-(T_k|t)$ can be similarly estimated. Given a target forward velocity $v^\text{d}$ and assuming symmetric gait, ALIP defines the desired pre-impact angular momentum as $L^\text{des} = z_0 v^\text{d}$. 
We would like $\hat{L}^-(T_{k+1}|t)^\text{des} = L^\text{des}$ for deadbeat controller or $\hat{L}^-(T_{k+1}|t)^\text{des} = \alpha (\hat{L}^-(T_{k}|t) - L^\text{des})$ where $\alpha \in [0,1)$ for a converging controller. Substituting either yields the foot placement law:
\begin{align*}
    \scalebox{1.05}{$u_x = \frac{\cosh(\lambda T) \hat{L}^-(T_k|t) + z_0\lambda \sinh (\lambda T )\hat{p}^-(T_k|t) - \hat{L}^-(T_{k+1}|t)^\text{des} }{ z_0\lambda \sinh (\lambda T )}$}
\end{align*}
\begin{remark}
The original ALIP formulation computes the swing foot location relative to the CoM~\cite{gong_angular_2021}. To maintain consistency across planners in RoMoCo, we express it as foot placement relative to the stance foot.
\end{remark}

\subsection{Hybrid(H)-LIP}
The H-LIP model~\cite{xiong_3-d_2022} extends ALIP by incorporating the double support (DS) phase. While the original formulation uses velocity as the second state, we retain angular momentum $L$ for consistency. During DS, the CoM is assumed to move at constant velocity, and thus the angular momentum remains constant. The resulting DS dynamics are:
\begin{equation}
\frac{d}{dt}
    \begin{bmatrix} p\\L\end{bmatrix} = \begin{bmatrix} 0 & \frac{1}{z_0}\\0 & 0\end{bmatrix}\begin{bmatrix} p\\L\end{bmatrix} \eqqcolon A_\text{DS} \begin{bmatrix} p\\L\end{bmatrix}.
\end{equation}
Walking dynamics alternate between SS and DS. The SS-to-DS transition is continuous, while the DS-to-SS transition includes the impact map. Over one step, the state evolves through the sequence
$\mathbf{x}_{\text{SS},k}^- = \mathbf{x}_{\text{DS},k}^+$ $\rightarrow$ $\mathbf{x}_{\text{DS},k}^-$  $\rightarrow$  $\mathbf{x}^{+}_{\text{SS},k+1}$ $\rightarrow$  
$\mathbf{x}^{-}_{\text{SS},k+1}$. Let $\mathbf{x}_{\text{SS},k}^-$ denote the state at the end of the $k$-th SS phase, the transition across one complete walking step is:
\begin{equation}
    \mathbf{x}^{-}_{\text{SS},k+1} 
    = e^{A_\text{SS} T_\text{SS}}(e^{A_\text{DS} T_\text{DS}}
      \mathbf{x}^{-}_{\text{SS},k} 
      + 
      B_\Delta u_k),
    \label{eq:transition}
\end{equation}
where $u_k$ is the $k$-th step size input, $T_\text{SS}$ and $T_\text{DS}$ are predefined gait time. From this point on, we treat the state at the end of the SS phase as the discrete state of the H-LIP, and drop unnecessary subscripts for clarity. The compact discrete-time dynamical system is:
\begin{equation}
    \mathbf{x}_{k+1} = A^H \mathbf{x}_k + B^H u_k,
    \label{eq:s2s_dynamics}
\end{equation}
where $A^H = e^{A_\text{SS}T_\text{SS} + A_\text{DS} T_\text{DS}}$ and $B^H = e^{A_\text{SS}T_\text{SS}}B_\Delta$
which is referred to as the S2S dynamics of the H-LIP. For a desired forward velocity $v^\text{d}$, the nominal step size is $u^* = v^\text{d}(T_\text{SS} + T_\text{DS})$, and the fixed point for states is solved by setting $\mathbf{x}_{k+1} = \mathbf{x}_{k} = \mathbf{x}^*$, which is $\mathbf{x}^* = (I-A^H)^{-1} B^H u^*$. We choose the following controller to stabilize the error dynamics between the robot and the H-LIP model:
\begin{align}
    u_\text{sw} = K (\hat{\mathbf{x}}^-(T_k|t) - \mathbf{x}^*) + u^*
\end{align}
where $K$ is the LQR or deadbeat gain and $\hat{\mathbf{x}}^-(T_k|t)$ is denotes the estimated pre-impact state at current time $t$, consistent with the ALIP formulation.

\subsection{Multi-domain(M)-LIP}
MLIP is designed for multi-domain heel-to-toe walking but also applies to flat-foot walking. It extends the LIP by adding the zero-moment point (ZMP) as an explicit state to capture foot-rollover effects. The ZMP is the point on the ground where the net moment of all contact forces is zero. The continuous-time dynamics of the MLIP are written as:
\begin{align}
    \dot{\bar{\mathbf{x}}} 
    &= A_{\text{ct}} \, \bar{\mathbf{x}} + B_{\text{ct}} \, \dot{p}_{\text{zmp}}, 
    \\
    \bar{\mathbf{x}} = 
    \begin{bmatrix} p \\ L \\ p_{\text{zmp}} \end{bmatrix}, \quad
    A_{\text{ct}} &= 
    \begin{bmatrix}
        0 & \tfrac{1}{z_0} & 0 \\
        g & 0 & -g \\
        0 & 0 & 0
    \end{bmatrix},
    \,
    B_{\text{ct}} =
    \begin{bmatrix} 0 \\ 0 \\ 1 \end{bmatrix},
    \label{eq:zlip_ct}
\end{align}
where $p_{\text{zmp}}$ is the horizontal ZMP location defined relative to the stance pivot.
For flat-footed walking, the stance pivot is placed directly under the ankle. Specifically, for a domain of duration $T_\ti$, the state at the end of that domain is:
\begin{equation}
    \bar{\mathbf{x}}_\ti^{-} 
    = e^{A_{\text{ct}} T_\ti} \, \bar{\mathbf{x}}_\ti^{+} 
      + \int_{0}^{T_\ti} e^{A_{\text{ct}}(T_\ti - t)} B_{\text{ct}} \, \dot{p}_{\text{zmp},\ti}(t) \, dt,
    \label{eq:zlip_solution}
\end{equation}
For SS, we consider $\dot{p}_{\text{zmp},i}(t) = 0$, and for DS, we assume  $\dot{p}_\text{zmp, DS}(t) = \frac{u_\text{sw}}{T_\text{DS}}$. Together with the impact solution for ZMP, where
\begin{align}
    p_\text{zmp, SS}^+ = p_\text{zmp, DS}^- - u_\text{sw},
\end{align}
we can obtain the step-to-step dynamics by evaluating the closed-form integral across domains and incorporating the impact maps. The same foot-placement feedback used for H-LIP then applies directly to the MLIP S2S dynamics.
\begin{remark}
Viewed through the ZMP formulation of the MLIP, the H-LIP assumption of constant CoM velocity during the DS phase is equivalent to constraining the ZMP to lie directly under the CoM. In this sense, H-LIP and flat-footed MLIP differ only in their assumptions on ZMP evolution during DS, while sharing the same SS dynamics.
\end{remark}
\subsection{Divergent Component of Motion (DCM)}
The eigenstructure of the LIP dynamics consists of a stable and an unstable manifold. The DCM corresponds to the unstable subspace of the LIP dynamics. It is defined as 
\begin{align*}
    \xi = p + \frac{\dot{p}}{\lambda} = p + \frac{L}{\lambda z_0}
\end{align*}
The transformation from the original state $(p,L)$ to the DCM coordinates is given by
\begin{align*}
    \begin{bmatrix}
        p \\
        \xi
    \end{bmatrix} = 
    \begin{bmatrix}
        1 & 0\\
        1 & \frac{1}{\lambda z_0} \\
    \end{bmatrix}
    \begin{bmatrix}
        p \\ L
    \end{bmatrix},
\end{align*}
which is an invertible linear transformation. When expressed at the dynamics level, this corresponds to a similarity transformation of the system matrix, which preserves eigenvalues and thus stability properties.
Substituting the LIP dynamics into this new coordinate system yields
\begin{align*}
\frac{d}{dt}
    \begin{bmatrix}
        p \\
        \xi
    \end{bmatrix} = \begin{bmatrix}
        -\lambda & \lambda \\
        0 & \lambda
    \end{bmatrix}
    \begin{bmatrix}
        p \\
        \xi
    \end{bmatrix}
\end{align*}
As the upper-triangular structure reveals, the CoM is exponentially attracted to the DCM, while the DCM itself evolves along the unstable manifold. 
Thus, regulating the DCM is sufficient to stabilize the overall LIP dynamics.

Early DCM formulations, similar to LIP, assumed fixed step time and location, with stabilization achieved through ZMP control~\cite{englsberger_three-dimensional_2013}. Later, \cite{khadiv_stepping_2016} introduced swing-foot adjustment based on DCM, and \cite{khadiv_walking_2020} proposed a QP-based planner that simultaneously optimized foot placement and step timing. The existing DCM formulation does not include DS. However, by augmenting the state with ZMP and adopting the same assumptions as MLIP for DS, one can derive a step-to-step map for DCM that incorporates double support:
\begin{align}
    \xi_{k+1} = a_\xi \xi_{k} + b_\xi u_k
\end{align}
where $a_\xi = e^{\lambda(T_{SS} + T_{DS})}$ and $b_\xi = -e^{\lambda T_{SS}} ( \frac{e^{\lambda(T_{DS}}-1}{\lambda T_{DS}} )$ if $T_{DS} > 0$ and $b_\xi = -e^{\lambda T_{SS}}$ if $T_{DS} = 0$. The fixed-point calculation and error-feedback controller then follow directly from the earlier derivations.

\begin{remark}The DCM and capture point (CP) have often been considered interchangeably in the literature. Mathematically, the DCM represents the unstable mode of the CoM dynamics and captures where the CoM “wants to go” if no action is taken. The CP is a physical interpretation of this quantity; it is the point on the ground where the robot must place its foot in order to bring itself to a stop. For the LIP model, the CP is equivalent to the DCM projected onto the ground plane.
\end{remark}


In summary, ALIP, H-LIP, MLIP, and DCM can all be interpreted as extensions of the canonical LIP model, each enriching the dynamics with additional structure. The presented unified perspective is the theoretical foundation that enables planners to be treated as interchangeable modules within the \texttt{romoco\_planner} package, allowing consistent embedding of foot-placement laws, straightforward switching between planners, and side-by-side evaluation of their relative performance. 


\section{Controller Output Embedding}
The bridge between the high-level ROM planner and the low-level torque controller is the \texttt{romoco\_output} package. Its purpose is twofold: first, to translate the discrete \texttt{PlannerOutput} into continuous, whole-body task objectives, and second, to formulate the holonomic constraints imposed by foot-ground contact and internal kinematic loops.

The whole-body controller's primary goal is to drive an error vector to zero. For a task i, it is defined as $\boldsymbol{y}_\ti \coloneqq \boldsymbol{y}_\ti^\text{a} - \boldsymbol{y}_\ti^\text{d}$, where $\boldsymbol{y}_\ti^\text{a}$ is the actual output from the robot's kinematics and $\boldsymbol{y}_\ti^\text{d}$ is the desired trajectory derived from the planner. For flat-footed walking, we define a output vector 
\begin{multline}
\y_\text{w} = [ z_\text{com}, \; \theta^z_{\text{st hip}}, \; \theta^y_{\text{pelvis}}, \; \theta^x_{\text{pelvis}}, \\
\{x,y,z\}_{\text{sw foot}}, \; \theta^z_{\text{sw hip}}, \; \{\theta^x, \theta^y\}_{\text{sw foot}} ].
\end{multline}
Here, $z_\text{com}$ represents the CoM height relative to the stance foot. $\theta^z_\text{st hip}$ and $\theta^z_\text{sw hip}$ are motor yaw angles to regulate body yaw orientation. $\theta^y_\text{pelvis}$ and $\theta^x_\text{pelvis}$ are computed from the torso relative displacement using the small-angle approximation, effectively providing torso pitch and roll angles in the local yaw frame. This avoids the need to extract Euler angles and yields numerically stable outputs for feedback. The $\{ x,y,z  \}_\text{sw foot}$
 denote the positions of the swing foot relative to the stance foot, expressed in the local yaw frame. Finally, $\{\theta^x, \theta^y\}_\text{sw foot}$, where the first is only used for robots with ankle roll motors, are computed in the same manner as the torso angles using small-angle approximations.
This set defines the regulation outputs during SS; in DS, only a subset is active, consistent with the contact holonomic constraints. 

\begin{figure}[t]
\centering
\vspace{2mm}
  \includegraphics[width=\linewidth]{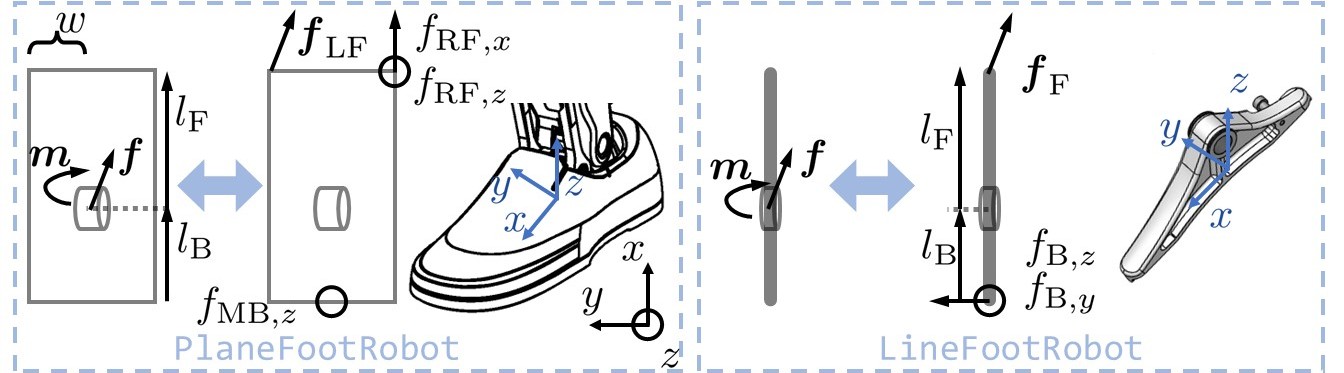}
  \caption{Visualization of holonomic contact constraints and equivalent wrench mappings using representative contact points. For plane-foot contact, forces at left-front (LF), right-front (RF), and mid-back (MB) are mapped to an equivalent wrench; for line-foot contact, constraining two points at the front and back suffice.}
  \vspace{-6mm}
  \label{fig::holonomic}
\end{figure}

To ensure physically consistent motion, the controller must enforce foot-ground contact constraints. Since Pinocchio directly provides linear Jacobians of frames but not explicit pitch/roll chains, we define contact by constraining multiple representative points on the foot sole (Fig. \ref{fig::holonomic}). For a planar contact, three non-collinear points are constrained; for a line contact, two points are used~\cite{caron_stability_2015}. 
This results in a holonomic Jacobian that constrains foot translation and rotation as required by the contact mode. This multi-point model also provides a consistent framework for handling contact forces.
As illustrated in Fig.~\ref{fig::holonomic}, we construct a linear transformation that maps multi-point forces into an equivalent 6D wrench (the line-foot case follows analogously):
\begin{align}
\scalebox{0.9}{$
\begin{bmatrix}
        f_x \\ f_y \\ f_z \\ m_x \\ m_y \\ m_z
    \end{bmatrix}
    = 
    \begin{bmatrix}
        1 & 0 & 0 & 1 & 0 & 0\\
        0 & 1 & 0 & 0 & 0 & 0\\
        0 & 0 & 1 & 0 & 1 & 1\\
        0 & 0 & w & 0 & -w & 0\\
        0 & 0 & l_\text{B} & 0 &l_\text{B} & l_\text{F}\\
        -w & -l_\text{B} & 0 & w & 0 & 0
    \end{bmatrix}
    \begin{bmatrix}
        f_{\text{LF},x} \\ f_{\text{LF},y} \\ f_{\text{LF},z} \\ f_{\text{RF},x} \\ f_{\text{RF},z} \\ f_{\text{MB},z}
    \end{bmatrix}
$} 
\end{align}
For three non-collinear points, the conversion matrix is full-rank, ensuring a well-defined mapping for plane-foot contact. 
Contact wrench feasibility is further enforced through friction-pyramid and ZMP constraints, applied in the original point-angle coordinates. This formulation allows RoMoCo to uniformly handle different contact modes (e.g., line or plane) in the whole-body controller while being compatible with Pinocchio-based kinematics and dynamics computations.

\begin{figure*}[t]
  \vspace{2mm}
\centering\includegraphics[width=0.95\textwidth]{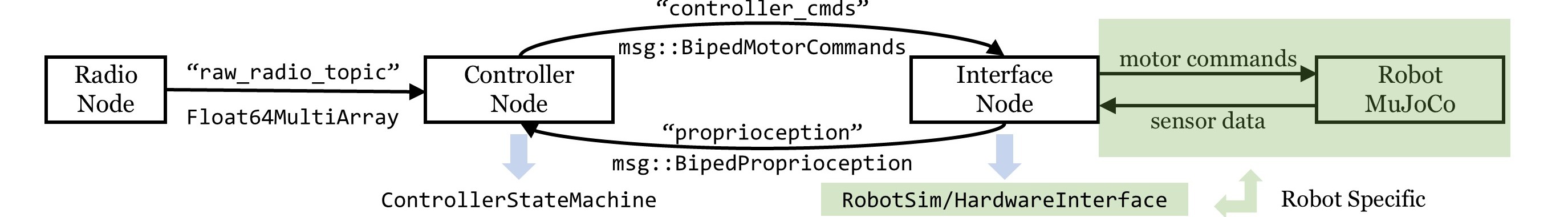}
  \caption{ROS2 overview of asynchronous simulation and hardware execution. 
The diagram shows the data flow between nodes, including topic names and message types. 
The same interface node supports both simulation and hardware via \texttt{RobotSim/HardwareInterface}.}
  \vspace{0mm}
  \label{fig::ros2_overview}
\end{figure*}

\section{Low Level Controllers}
Whole body controllers close the loop between the embedded planner outputs and the robot actuators. Given an output vector $\y_\ti(\q)$, holonomic constraints $J_\ti(\q)$, the controller produces joint torques $\boldsymbol{\tau}$ that minimize output error while respecting holonomic constraints. The \texttt{romoco\_control} package exposes several interchangeable solver backends, allowing users to select the suitable method based on computational budget, hardware platform, and research goals.

The robot's continuous dynamics is governed by the Euler-Lagrange equations, assuming non-slip contact conditions. The equation of motion is given by:
\begin{align}
    &D(\q)\ddot{\q} + H(\q,\dot{\q}) = B \boldsymbol{\tau} + J_\ti(\q)^\intercal \boldsymbol{f}_\ti ,\label{eq::eom} \\ 
    &J_\ti(\q)\ddot{\q} + \dot{J}_\ti(\q,\dot{\q})\dot{\q} = 0, \label{eq::hol} 
\end{align}
where $\boldsymbol{q}\in Q$ is a set of generalized coordinates in the $n$-dimensional configuration space $Q$, $D(\q)\in \R^{n \times n}$, $H(\q,\dot{\q}) \in \R^{n}$, $B\in \R^{n\times m}$ are the inertia matrix, the collection of centrifugal, Coriolis and gravitational forces, and the actuation matrix, respectively. The input torque is denoted by $\boldsymbol{\tau}\in U \subseteq \R^m$. $J_\ti(\q)\in\R^{n\times h_\ti}$ is the task and domain-specific Jacobian matrix related to contact constraints, and $\boldsymbol{f}_\ti\in\R^{h_\ti}$ represents the corresponding constraint wrench. 
\subsection{Task-Space Control Quadratic Programming (TSC-QP)}
For robots with accurate dynamics models and torque-controlled actuators, TSC-QP~\cite{bouyarmane_quadratic_2019} is the most powerful method as it optimizes for task accelerations while explicitly respecting dynamics, contact forces, and torque limits. At each control loop for task i, the following QP is formulated in \texttt{TorqueSolverTSCQP}: 
\begin{align}
   \underset{\boldsymbol{\ddot{q}}, \boldsymbol{\tau}, \boldsymbol{f}_\ti } {\text{min}}  & \quad ||\ddot{\y}^\text{a}_\ti(\q,\dot{\q},\ddot{\q}) - \ddot{\y}^\text{d}_\ti - \ddot{\y}^\text{t}_\ti ||^2_{Q_\ti}, \label{eq::TSC} \tag{TSC-QP} \\
\text{s.t.}  & \quad   \text{Eqs.}~\eqref{eq::eom}, \eqref{eq::hol},  \tag{Dynamics} \\
 & \quad    A_{\text{GRF},\ti} \boldsymbol{f}_\ti  \leq \boldsymbol{b}_{\text{GRF},\ti}, \tag{Contact} \\ 
 & \quad  \boldsymbol{\tau}_{lb} \leq \boldsymbol{\tau} \leq \boldsymbol{\tau}_{ub}.  \tag{Torque Limit}
\end{align}
Here, $Q_\ti$ denotes a weight matrix, $\ddot{\y}^\text{a}_\ti$ and $\ddot{\y}^\text{d}_\ti$ represent the actual and desired accelerations of the output $\y_\ti$, 
and $ \ddot{\y}^\text{t}_\ti = - K_p \y_\ti - K_d \dot{\y}_\ti$ is the target acceleration that enables exponential tracking, with $K_p, K_d$ being the proportional and derivative gains.
The affine contact constraint on $\boldsymbol{f}_\ti$ enforces the contact friction pyramid and ZMP bounds given the contact condition of each domain, while $\boldsymbol{\tau}_{lb}$ and $\boldsymbol{\tau}_{ub}$ are torque limits. Solving the optimization yields the optimal torque $\boldsymbol{\tau}$ for the robot. The QP is solved using Clarabel~\cite{Clarabel_2024}, which we found to provide a good balance of accuracy and computational speed.

\subsection{Inverse Dynamics (ID)}
The inverse dynamics controller provides a computationally lighter alternative to TSC–QP by enforcing only equality constraints, making it faster but less robust to constraint violations. Conceptually, it can be viewed as a second-order controller equivalent to a TSC–QP formulation without inequality constraints, where the optimization cost is treated as an equality condition. To improve numerical stability, we adopt the projection method~\cite{park_contact_2006}, which reformulates the dynamics \eqref{eq::eom}, \eqref{eq::hol} as constrained dynamics in operational space coordinates. An implementation is provided in the \texttt{TorqueSolverInvDyn} class.

\subsection{Inverse Kinematics (IK)}
For robots with high-geared actuators or embedded position controllers, kinematics-based methods are often more practical than torque optimization. IK is commonly used to generate reference trajectories or enforce task-space constraints such as end-effector positions. 
The core idea is to solve for the target motor positions $\q^{\text{d}}_m$ and velocities $\dot{\q}^{\text{d}}_m$ while respecting the holonomic constraints
, which are then tracked with a joint-space PD controller:
\begin{equation}
    \boldsymbol{\tau}_{\text{fb}} =
    K_{p}^m\left( \mathbf{q}^{\text{d}}_m - \mathbf{q}_m^\text{a} \right) +
    K_d^m\left( \dot{\mathbf{q}}^{\text{d}}_m - \dot{\mathbf{q}}_m^\text{a} \right),
    \label{eq:pd_torque}
\end{equation}
where $K_p^m, K_d^m \in \mathbb{R}^{m \times m}$ are diagonal gain matrices.
A feedforward gravity compensation term, $\boldsymbol{\tau}_{\text{ff}} = \mathbf{g}_\tau(\mathbf{q})$, is added, which is essential for stance support and equivalent to solving an ID problem with zero target acceleration \cite{mistry_inverse_2010}.
The final controller for the torque-controlled robot is given by
$\boldsymbol{\tau} = \boldsymbol{\tau}_{\text{ff}} + \boldsymbol{\tau}_{\text{fb}}$. 
For motors with embedded position loops, $K_p^m$, $K_d^m$, $\q_m^\text{d}$, $\dot{\q}_m^\text{d}$, and $\boldsymbol{\tau}_\text{ff}$ are sent directly as references.

In legged robots, holonomic constraints such as foot contacts or closed kinematic chains must be enforced. For task i, we ensure consistency by projecting task-space motions into the constraint-null space using $N_\ti = I - J_\ti^{\dagger} J_\ti$, where ${(\cdot)}^\dagger$ denotes the Moore-Penrose pseudoinverse. The active task Jacobian consistent with holonomic constraints is:
\begin{equation}
    \bar{J}_y = J_y\,N_\ti.
\end{equation}

\subsubsection{Velocity-Based Inverse Kinematics (VEL-IK)}
Given desired outputs $\y^\text{d}$, $\dot{\y}^\text{d}$ and measured output and joint angles $\y^\text{a}$, $\q^\text{a}$, the IK updates are obtained from:
\begin{align}
    \q^\text{d} &= \q^\text{a} + \bar{J}_y^{\dagger} 
    \left( \y^\text{d} - \y^\text{a} \right),
    \label{eq:ik_position}\\
    \dot{\q}^\text{d} &= \bar{J}_y^{\dagger}
    \,\dot{\y}^\text{d},
    \label{eq:ik_velocity}
\end{align}
We select the motored indices of (\ref{eq:ik_position})--(\ref{eq:ik_velocity}) as the final motor target position and velocities, $\q^\text{d}_m$, $\dot{\q}^\text{d}_m \in \R^m$.

\subsubsection{Position-Based Inverse Kinematics (POS-IK)}
In addition to VEL-IK, the toolbox provides a position-based IK solver formulated as an iterative Newton–Raphson method. Assuming $\q^\text{d}_0 = \q^\text{a}$, the update rule for $\q$ is:
\begin{equation}
    \q^\text{d}_{k+1} = \q^\text{d}_{k} + \bar{J}_y(\q_k)^\dagger\left(\y^\text{d} - \y^\text{a}(\q_k^\text{d})\right),
\end{equation}
until $\|\y^\text{d} - \y^\text{a}(\q_k^\text{d})\| < \epsilon$ or $k \geq k_{\max}$.
After convergence, a VEL-IK step is applied to compute $\dot{\q}^\text{d}$ consistent with holonomic constraints. 

\section{Interface}
With the controller modules defined, we next describe the interface layer that enables deployment on different robots and environments.
A central goal of RoMoCo is to make controllers portable across different robots and deployment environments. The \texttt{InterfaceBase} class provides this abstraction by exposing a unified API for proprioception, actuation, and communication with both simulators and hardware. We recommend a staged workflow, beginning with in-air testing, progressing to standing, and finally deploying walking behaviors. To support this process, the toolbox includes ready-to-use stacks for robot Cassie, G1, and H1.

For simulation, RoMoCo provides a generic MuJoCo interface via the \texttt{romoco\_mujoco} package, as well as robot-specific interfaces when custom features are needed, see \texttt{cassie\_stack} for example \cite{cassie_mujocosim}. The simulation interface supports both (i) \emph{synchronous simulation}, where the controller and simulator run in the same thread for deterministic debugging, and (ii) \emph{asynchronous simulation}, where the controller and simulator run in separate ROS nodes, better matching the hardware deployment conditions. Both modes can use either ground-truth simulated velocities or velocities estimated from a Kalman filter, allowing users to test estimation sensitivity before moving to hardware.
Fig. \ref{fig::ros2_overview} provides a reference to the implemented ROS2 nodes and communication between nodes for the asynchronous simulation and hardware communication. 

On hardware, the interface layer bridges robot-specific communication protocols with RoMoCo’s unified internal messages (\texttt{BipedProprioception} and \texttt{BipedMotorCommands}). For example, Cassie integrates Agility’s UDP backend, while G1 and H1 use DDS. Despite protocol differences, all controllers interact with the same upstream API, ensuring portability across platforms. The table below summarizes the tested robots and controllers currently supported in RoMoCo. 
\textbf{\begin{table}[h!]
\centering
\begin{tabular}{|c|c|c|c|c|c|}
\hline
\textbf{Robot} & 
\multicolumn{2}{c|}{\textbf{Cassie}} & 
\multicolumn{2}{c|}{\textbf{G1}} & 
\textbf{H1} \\ \hline
 Controller & Sim & Hardware & Sim & Hardware & Sim \\ \hline
Vel IK / Pos IK & \checkmark &  & \checkmark & \checkmark & \checkmark \\ \hline
QP / ID & \checkmark & \checkmark & \checkmark &  & \checkmark \\ \hline
\end{tabular}
\vspace{-2mm}
\end{table}}

\subsection{Case Study: Cassie}
Cassie, developed by Agility Robotics, is a 3D underactuated biped with line feet and internal six-bar linkages that impose kinematic loop constraints. Assuming rigid springs, each leg has five actuated joints and one passive tarsus, with a floating-base pelvis contributing six additional DOF for a total of 18. Within RoMoCo, both internal holonomic constraints (six-bar linkages) and external ones (foot–ground contacts) are enforced in the robot model and passed to the low-level controller as holonomic constraints, ensuring that the generated torques remain consistent with Cassie’s constrained dynamics.

\begin{figure}[t]
\vspace{2mm}
  \includegraphics[width=\linewidth]{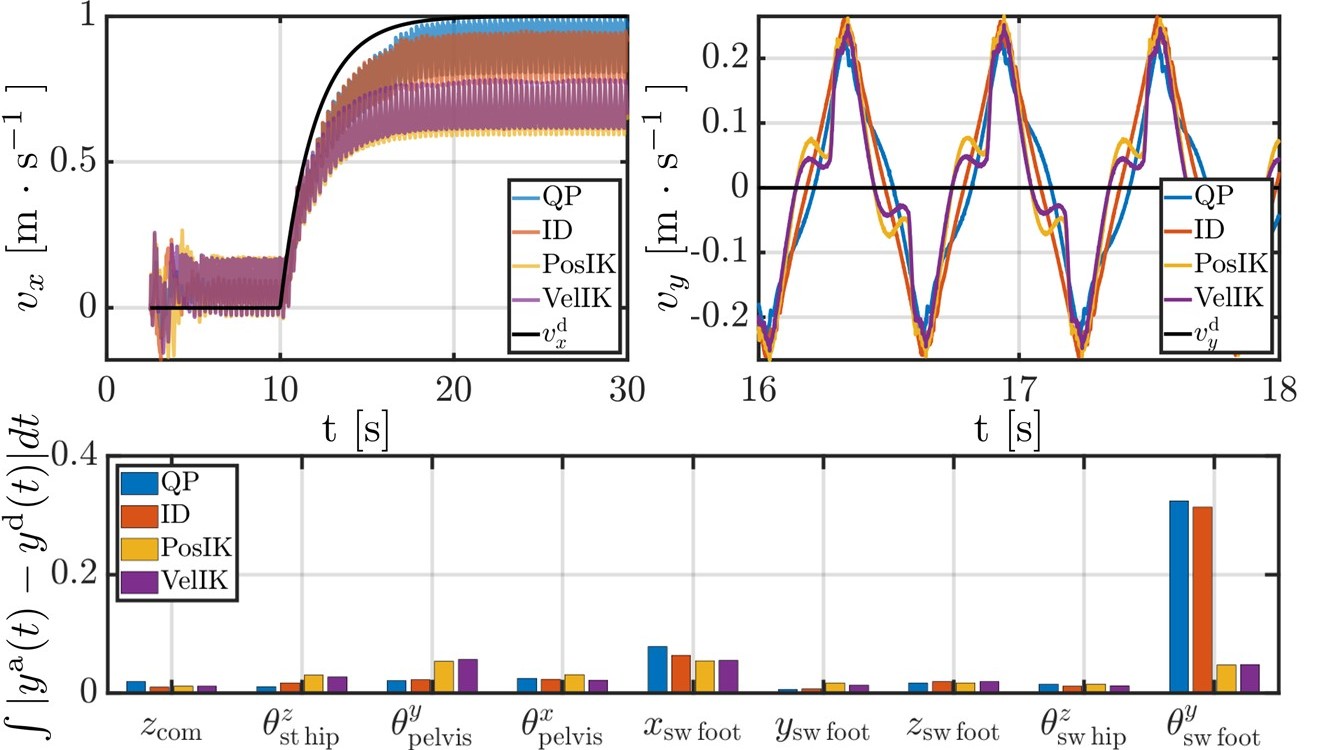}
  \caption{Tracking performance of four control formulations. Top left: forward velocity tracking; top right: lateral velocity tracking; bottom: integrated output tracking error from $t=$28s to $t=$30s.}
  \vspace{-4mm}
  \label{fig::cassie_controller_comparison}
\end{figure}

To evaluate controller performance, we compared four controller formulations on Cassie: QP, ID, POK-IK, and VEL-IK for the DCM planner. Fig.~\ref{fig::cassie_controller_comparison} illustrates representative results. The top panels show tracking of forward and lateral velocities, and the bottom panel summarizes integrated output tracking errors across multiple tasks. QP achieves the most consistent velocity tracking because it tracks the output while satisfying contact and actuation constraints. ID performs similarly but is more sensitive to modeling errors and produces larger deviations in certain tasks. In contrast, IK methods prioritize end-effector tracking: they achieve superior swing foot orientation control but at the expense of overall velocity regulation and pelvis stability. 
Overall, dynamics-based controllers are advantageous for stance regulation, while IK methods exploit their simplicity and strong end-effector tracking when joint-level loops dominate. 
RoMoCo’s modular design allows users to define custom controllers for new robots while still benchmarking them against established approaches.

\begin{figure}[t]
\vspace{2mm}
  \includegraphics[width=\linewidth]{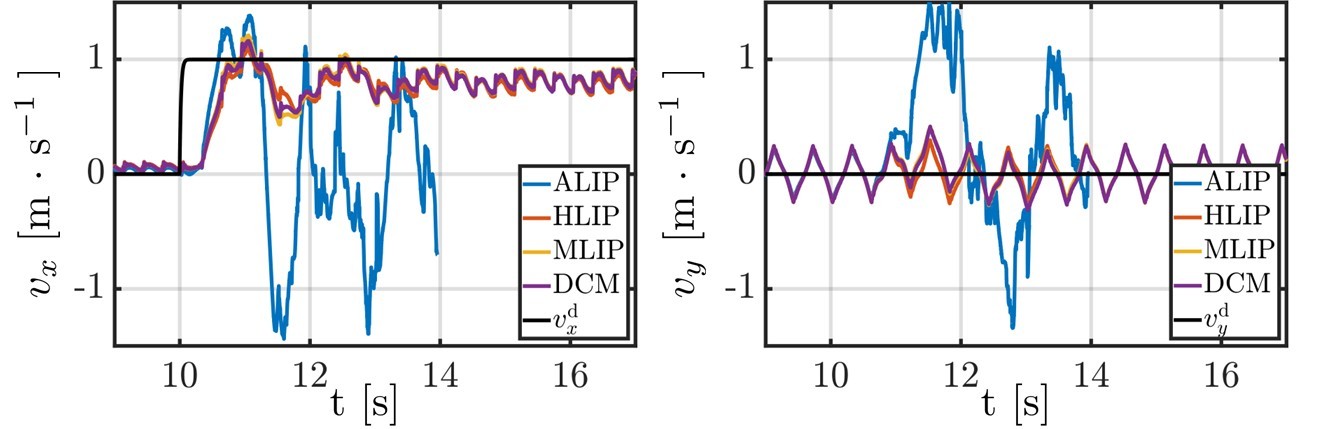}
  \vspace{-2mm}
  \caption{Performance of four ROM planners under aggressive velocity commands: ALIP, HLIP, MLIP, and DCM. The plots show the commanded forward velocity
 (left) and lateral velocity (right). HLIP, MLIP, and DCM achieve stable tracking, while ALIP exhibits loss of stability.}
  \vspace{-6mm}
  \label{fig::cassie_planner_comparison}
\end{figure}
We further compared four ROM planners on Cassie: ALIP, H-LIP, MLIP, and DCM, with the QP controller. Fig.~\ref{fig::cassie_planner_comparison} shows tracking performance for commanded forward and lateral velocities. For aggressive speed commands, ALIP exhibited large oscillations that quickly led to instability, reflecting its limitations from neglecting error dynamics between the robot and the model. In contrast, H-LIP, MLIP, and DCM maintained walking with similar tracking performance. Since ALIP does not support a double-support phase, the comparison is restricted to single-support dynamics, under which the remaining three planners perform nearly identically.

The HLIP planner with the QP and ID controllers was successfully deployed on Cassie hardware prior to refactoring the codebase into the robot-generic toolbox presented here. Although the Cassie platform is currently unavailable and these experiments cannot be reproduced, we still provide the hardware interface in the toolbox. This allows users with access to Cassie to directly run RoMoCo controllers, and also serves as a concrete example of integrating a custom UDP-based communication backend into our interface layer.

\subsection{Case Study: G1}
The G1 humanoid robot \cite{g1}, developed by Unitree Robotics, is modeled with flat-foot contact and full actuated joints. Each leg has six actuators, and together with the waist and the arms, the robot has 35 DOF. In the current release, the upper-body joints are fixed at nominal angles using a PD controller, while locomotion control focuses on the lower body. This setup highlights how RoMoCo’s modular design accommodates robots with flat-foot assumptions, while allowing specific DOFs to be isolated for control.

In simulation, G1 uses the generic \texttt{romoco\_mujoco}, while on hardware it communicates through DDS~\cite{unitree_sdk}.
Unlike Cassie’s torque-controlled motors, G1’s actuators include embedded position and velocity loops with feedforward torque. When commanded with torque alone, the system exhibits significant chatter. Thus, G1 provides a contrasting case study of a fully actuated humanoid with high-gain position control, where IK-based methods are better suited because they exploit the built-in loops for stable tracking.

\begin{figure}[t]
\vspace{2mm}
  \includegraphics[width=\linewidth]{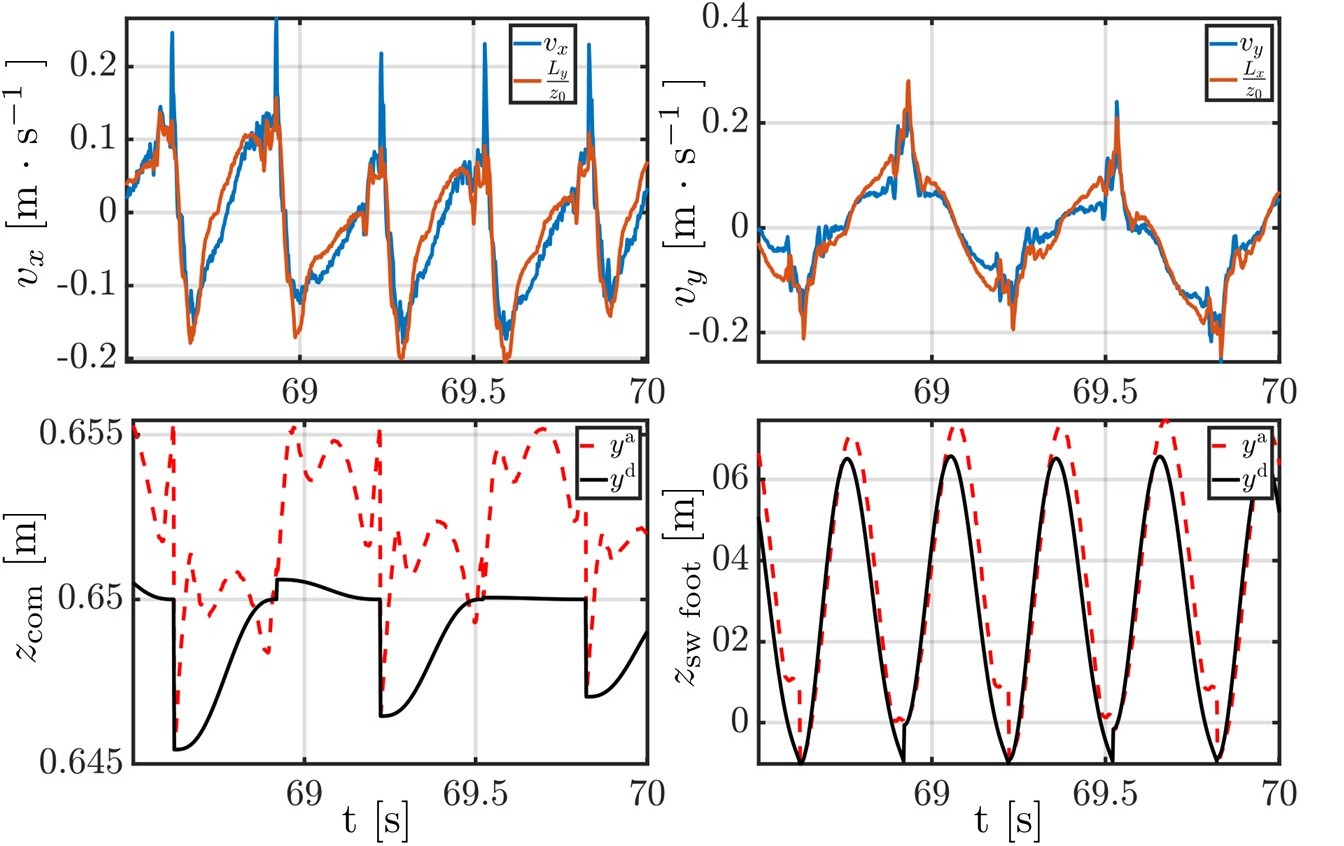}
  \caption{G1 Hardware CoM and output tracking: the top row illustrates correlations between CoM velocity and angular momentum about stance pivot, while the bottom row shows tracking of $z_\text{com}$ and $z_\text{sw}$.}
  \vspace{-6mm}
  \label{fig::g1_hardware}
\end{figure}

Figure~\ref{fig::g1_hardware} shows representative experimental results using the DCM planner with a POS-IK controller when commanded to step in place. The top row presents the linear velocity and angular momentum about the stance pivot, with the latter exhibiting smoother variations as expected. The bottom row reports tracking performance for vertical CoM position and swing foot height, where the desired outputs (black) are closely followed by the measured trajectories (red). These results confirm that the same reduced-order planning and control pipeline can be deployed on hardware with consistent tracking of key locomotion outputs.

\section{Discussion}
RoMoCo provides a unified platform for integrating reduced-order locomotion planners and whole-body controllers, lowering the barrier to comparative studies in legged robotics. By decoupling planning, control, and simulation through a modular API, RoMoCo allows researchers to evaluate different algorithmic components in isolation or in combination. The same walking task can be executed with different reduced-order models and torque solvers, configured through lightweight YAML specifications. 

Despite these advantages, RoMoCo currently has limitations. Its primary backend is MuJoCo, which is efficient and widely adopted but limits large-scale GPU-accelerated training compared to frameworks such as IsaacGym. Future work will focus on supporting additional physics engines, incorporating data-driven controllers alongside the model-based methods currently emphasized, and a better data pipeline so it can provide tighter integration with RL.

\section{Conclusion}

We have presented RoMoCo, an open-source software library for the development of reduced-order model-based planners and whole-body controllers for bipedal and humanoid robots. RoMoCo combines modularity, extensibility, and reproducibility within a unified platform, enabling rapid prototyping and fair comparison of locomotion strategies. 
We envision RoMoCo serving as a community resource that accelerates progress in legged robotics by fostering transparent, reproducible, and extensible research.




\bibliographystyle{IEEEtran}
\balance

\bibliography{references,misc}

\end{document}